\documentclass[11pt]{article}

\usepackage{amsmath,amssymb}
\usepackage{booktabs}
\usepackage{geometry}
\usepackage{enumitem}
\usepackage[numbers,sort&compress]{natbib}
\usepackage{hyperref}
\usepackage{tikz}

\usetikzlibrary{positioning,arrows.meta,decorations.pathreplacing}

\title{\bf Is Monte Carlo Tree Search Just Every-Visit Monte Carlo Control?}

\author{Xianyi Wu, ECNU}

\begin{document}

\maketitle

\begin{abstract}

Monte Carlo Tree Search (MCTS) and every-visit Monte Carlo (MC)
control are usually presented as different methods. MCTS is described
in the language of search---selection, expansion, simulation, and
backup---whereas MC control is described in the language of
reinforcement learning---trajectory sampling, return estimation,
action-value updating, and policy improvement.

This note argues that the difference is largely one of description.
At the level of trajectory generation and action-value updating, MCTS
can be viewed directly as every-visit Monte Carlo control. On states
that have already been visited, actions are selected according to the
current policy constructed from the available value information. On
states for which such information has not yet been acquired, the
current policy simply remains at its initial or default value. In MCTS
terminology, these two parts of the same evolving policy are called
the tree policy and the rollout policy, respectively. Expansion is the
first visit to a previously unrepresented state or action, while
backup is the ordinary every-visit Monte Carlo update.

Under this interpretation, the familiar four stages of MCTS reduce to
two basic operations:
\[
\text{trajectory sampling}
\quad+\quad
\text{every-visit Monte Carlo updating}.
\]
The tree provides a useful representation of the visited region and a
natural visualization of the search process, while UCB/UCT is one
possible rule for action selection. Neither changes the underlying
Monte Carlo control mechanism.

The purpose of this note is expository: to make this equivalence
explicit and to provide a simple unified interpretation of two methods
that are often taught and discussed separately.
\end{abstract}

\section{Introduction}

Monte Carlo (MC) control and Monte Carlo Tree Search (MCTS) are usually
introduced as different algorithms.

In reinforcement learning, every-visit MC control is described as a
procedure in which trajectories are sampled, Monte Carlo returns are
used to estimate action values, and the policy is progressively
modified using the resulting estimates
\cite{sutton2018reinforcement}.

MCTS, in contrast, developed primarily in the planning and game-search
literature \cite{coulom2006efficient,kocsis2006bandit,browne2012survey}.
Its basic operation is commonly described through four stages:
\[
\text{Selection}
\longrightarrow
\text{Expansion}
\longrightarrow
\text{Simulation}
\longrightarrow
\text{Backup}.
\]

This difference in terminology makes the two methods appear more
different than they actually are.

The close relationship between MCTS and reinforcement learning is not
new. In particular, Vodopivec, Samothrakis, and \v{S}ter
\cite{vodopivec2017mcts} gave a detailed treatment of the relationship
between MCTS and RL and advocated a unified view of learning, planning,
and search. Related connections between Monte Carlo planning,
adaptive sampling, and MDP solution methods have also appeared in the
simulation and planning literature
\cite{chang2005adaptive,fu2018mcts}.

The purpose of the present note is simpler and mainly expository. We
ask what MCTS looks like if its usual search terminology is translated
directly into the language of every-visit MC control.

The answer is remarkably simple.

Both methods repeatedly perform two basic operations:
\[
\boxed{
\text{sample a trajectory under the current policy}
}
\]
and
\[
\boxed{
\text{update the visited action values using Monte Carlo returns}.
}
\]

The main source of apparent difference is that MCTS assigns separate
names to different portions of the trajectory and explicitly
represents the region that has already been explored.

Once these terminological differences are removed, the familiar MCTS
operations admit the following interpretation:
\[
\begin{aligned}
\text{Selection}
&\leftrightarrow
\text{sampling on states where the policy has been updated},\\
\text{Expansion}
&\leftrightarrow
\text{first visit / initialization},\\
\text{Simulation}
&\leftrightarrow
\text{sampling on states still using the initial policy},\\
\text{Backup}
&\leftrightarrow
\text{every-visit Monte Carlo updating}.
\end{aligned}
\]

This leads to the central viewpoint of this note:
\[
\boxed{
\text{MCTS}
=
\text{Every-Visit Monte Carlo Control}
}
\]
at the level of trajectory sampling and Monte Carlo action-value
updating.

The remainder of the note explains this correspondence.

\section{Every-Visit Monte Carlo Control}

Consider an episodic Markov decision process and a sampled trajectory
\[
\tau
=
(S_0,A_0,R_1,S_1,A_1,\ldots,S_T).
\]

For a state--action pair visited at time $t$, define the Monte Carlo
return
\[
G_t
=
\sum_{k=0}^{T-t-1}
\gamma^k R_{t+k+1}.
\]

Every-visit Monte Carlo estimation updates $Q(S_t,A_t)$ every time
$(S_t,A_t)$ appears in the trajectory.

Let $N(s,a)$ denote the number of times $(s,a)$ has previously been
updated. The sample-average recursion is
\[
Q(s,a)
\leftarrow
Q(s,a)
+
\frac{1}{N(s,a)+1}
\left[
G_t-Q(s,a)
\right],
\]
followed by
\[
N(s,a)\leftarrow N(s,a)+1.
\]

Equivalently,
\[
Q(s,a)
\leftarrow
\frac{
N(s,a)Q(s,a)+G_t
}{
N(s,a)+1
}.
\]

Thus, a single trajectory supplies Monte Carlo observations for all
state--action pairs appearing along that trajectory.

MC \emph{control} adds action selection and policy improvement. In a
generic form, the current policy may be written as
\[
\pi_n(a\mid s)
=
\Phi\!\left(
Q_n(s,\cdot),N_n(s,\cdot)
\right),
\]
where $\Phi$ may be an $\epsilon$-greedy rule, a softmax rule, a UCB
rule, or another exploration mechanism.

The essential loop is therefore
\[
\boxed{
\text{current policy}
\rightarrow
\text{trajectory}
\rightarrow
\text{return}
\rightarrow
Q\text{-update}
\rightarrow
\text{updated policy}.
}
\]

\section{The MCTS Backup Is an Every-Visit MC Update}

Consider one MCTS simulation producing
\[
(S_0,A_0),(S_1,A_1),\ldots,(S_{T-1},A_{T-1})
\]
and corresponding returns $G_t$.

A common MCTS implementation maintains
\[
N(s,a)
\qquad\text{and}\qquad
W(s,a),
\]
where $N(s,a)$ is the visit count and $W(s,a)$ is the accumulated
return. The empirical action value is
\[
Q(s,a)=\frac{W(s,a)}{N(s,a)}.
\]

After another simulated return $G_t$,
\[
N(s,a)\leftarrow N(s,a)+1,
\]
\[
W(s,a)\leftarrow W(s,a)+G_t.
\]

Consequently,
\[
Q(s,a)
\leftarrow
Q(s,a)
+
\frac{1}{N(s,a)}
\left[
G_t-Q(s,a)
\right],
\]
where $N(s,a)$ now denotes the updated visit count.

This is exactly the sample-average every-visit Monte Carlo update.

Therefore,
\[
\boxed{
\text{MCTS backup}
=
\text{every-visit MC update}.
}
\]

The word ``backup'' describes the direction in which the information
is propagated through the stored search representation. It does not
introduce a different Monte Carlo estimator.

\section{One Policy, Not Two}

The most important apparent difference between MCTS and ordinary MC
control is the usual distinction between a \emph{tree policy} and a
\emph{rollout policy}.

This distinction can be expressed more simply.

Let
\[
\mathcal V_n
=
\left\{
s:
\text{action-value information has already been acquired at }s
\right\}.
\]

For
\[
s\in\mathcal V_n,
\]
the algorithm has learned quantities such as
\[
Q_n(s,a),\qquad N_n(s,a),
\]
and can therefore select actions according to an updated rule:
\[
\pi_n(a\mid s)
=
\Phi\!\left(
Q_n(s,\cdot),N_n(s,\cdot)
\right).
\]

Now consider
\[
s\notin\mathcal V_n.
\]

No action-value information has yet been acquired at this state.
Consequently, there is no learned information with which to modify
the initial action-selection rule. The policy at such a state remains
\[
\pi_0(a\mid s).
\]

The current policy over the entire state space can therefore be written
as
\[
\boxed{
\pi_n(a\mid s)
=
\begin{cases}
\Phi\!\left(Q_n(s,\cdot),N_n(s,\cdot)\right),
&
s\in\mathcal V_n,
\\[2mm]
\pi_0(a\mid s),
&
s\notin\mathcal V_n.
\end{cases}
}
\tag{1}
\]

Equation (1) removes the need to regard the tree policy and rollout
policy as fundamentally different policy objects.

In MCTS terminology,
\[
\boxed{
\text{tree policy}
=
\text{current policy on the learned region},
}
\]
whereas
\[
\boxed{
\text{rollout policy}
=
\text{initial part of the current policy on the unlearned region}.
}
\]

The policy is one evolving object. It has simply been updated on some
states and not yet updated on others.

\section{Selection and Rollout Are One Sampling Procedure}

Suppose one simulation produces
\[
s_0
\rightarrow
s_1
\rightarrow
s_2
\rightarrow
s_3
\rightarrow
s_4
\rightarrow
\cdots
\rightarrow
s_T.
\]

Assume that
\[
s_0,s_1,s_2\in\mathcal V_n,
\]
while
\[
s_3,s_4,\ldots\notin\mathcal V_n.
\]

At the first states, actions are sampled according to
\[
\Phi(Q_n,N_n).
\]
MCTS calls this part of the trajectory \emph{selection}.

At states for which no learned action-value information exists,
actions are sampled according to
\[
\pi_0.
\]
MCTS calls this part \emph{simulation} or \emph{rollout}.

But according to (1), both are simply portions of a trajectory sampled
under the same globally defined current policy $\pi_n$.

Thus,
\[
\boxed{
\text{Selection + Rollout}
=
\text{trajectory sampling under }\pi_n.
}
\]

Figure~\ref{fig:unified} illustrates this interpretation.
 \usetikzlibrary{calc}

\begin{figure}[t]
\centering

\begin{tikzpicture}[
    >=Latex,
    node distance=1.25cm,
    state/.style={
        circle,
        draw,
        minimum size=8mm,
        inner sep=1pt
    },
    lab/.style={
        font=\small,
        align=center
    }
]

\node[state] (s0) {$s_0$};
\node[state, right=of s0] (s1) {$s_1$};
\node[state, right=of s1] (s2) {$s_2$};
\node[state, right=of s2] (s3) {$s_3$};
\node[state, right=of s3] (s4) {$s_4$};
\node[state, right=of s4] (st) {$s_T$};

\draw[->] (s0) -- node[above] {$a_0$} (s1);
\draw[->] (s1) -- node[above] {$a_1$} (s2);
\draw[->] (s2) -- node[above] {$a_2$} (s3);
\draw[->] (s3) -- node[above] {$a_3$} (s4);
\draw[->] (s4) -- node[above] {$\cdots$} (st);

\draw[
    decorate,
    decoration={brace,amplitude=6pt},
    thick
]
($(s0.north west)+(0,0.55)$)
--
($(st.north east)+(0,0.55)$)
node[midway,above=9pt,lab]
{
One trajectory sampled under the current policy $\pi_n$
};

\draw[
    decorate,
    decoration={brace,mirror,amplitude=5pt},
    thick
]
($(s0.south west)+(0,-0.35)$)
--
($(s2.south east)+(0,-0.35)$)
node[midway,below=8pt,lab]
{
Selection\\
$\pi_n=\Phi(Q_n,N_n)$
};

\draw[
    decorate,
    decoration={brace,mirror,amplitude=5pt},
    thick
]
($(s3.south west)+(0,-0.35)$)
--
($(st.south east)+(0,-0.35)$)
node[midway,below=8pt,lab]
{
Rollout / simulation\\
$\pi_n=\pi_0$
};

\node[below=1.8cm of s3,lab] (exp)
{
Expansion\\
first visit / initialization
};

\draw[->,dashed] (exp.north) -- (s3.south);

\end{tikzpicture}

\caption{
The tree-policy and rollout-policy portions of MCTS can be interpreted
as two regions of a single evolving policy. On states with learned
action-value information, the current policy uses that information.
On states not yet learned, it retains its initial value $\pi_0$.
Selection, expansion, and rollout therefore form one trajectory-sampling
procedure.
}
\label{fig:unified}
\end{figure}
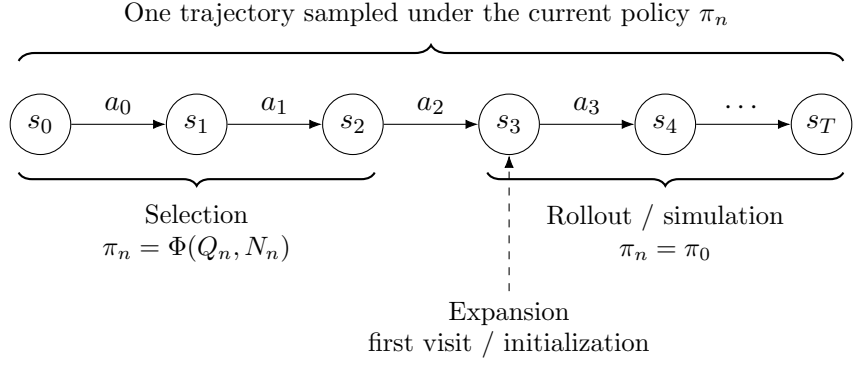

\section{Expansion Is First Visit}

MCTS assigns a separate name to the first encounter with a previously
unrepresented state or action: \emph{expansion}.

From the perspective of MC control, nothing statistically new happens
at this point.

A tabular MC-control description may conceptually initialize
\[
Q_0(s,a),\qquad N_0(s,a)
\]
for all state--action pairs.

There is no requirement, however, that all such entries be physically
created in advance. They can instead be created when first encountered.

Thus,
\[
\boxed{
\text{Expansion}
=
\text{first visit plus lazy initialization}.
}
\]

Whether the data structure is allocated in advance or expanded on
demand does not alter the underlying sampling or value-update rule.

\section{What Does the Tree Add?}

The tree in MCTS is extremely useful.

It records which portions of the decision process have been visited,
organizes parent--child relationships, stores visit counts and value
estimates, and makes the boundary between explored and unexplored
regions explicit.

It is also an excellent visualization device.

For the value-learning mechanism considered here, however, the
essential stored statistics are simply
\[
(s,a)
\mapsto
\bigl(Q(s,a),N(s,a)\bigr).
\]

These statistics could in principle be stored in a tree, a table, a
hash map, or another suitable data structure.

The explicit tree is particularly natural because MCTS usually
concentrates computation on trajectories originating from one current
state. But the use of a tree does not change either

\[
\text{how trajectories are sampled}
\]
or
\[
\text{how Monte Carlo returns update }Q.
\]

In this sense, the tree is a representation of the computation rather
than a different Monte Carlo learning principle.

\section{UCB Is an Action-Selection Rule}

UCT \cite{kocsis2006bandit} is probably the best-known instance of
MCTS.

A typical UCB selection rule is
\[
A
=
\arg\max_a
\left\{
Q(s,a)
+
c
\sqrt{
\frac{\log N(s)}
{N(s,a)}
}
\right\}.
\]

This rule determines how actions are selected in states where search
statistics are available.

But nothing about the Monte Carlo update requires UCB.

One could instead use $\epsilon$-greedy:
\[
A=
\begin{cases}
\arg\max_a Q(s,a),
&
\text{with probability }1-\epsilon,
\\
\text{an exploratory action},
&
\text{with probability }\epsilon,
\end{cases}
\]
or softmax:
\[
\pi_n(a\mid s)
=
\frac{
\exp\{Q_n(s,a)/\tau\}
}{
\sum_b\exp\{Q_n(s,b)/\tau\}
}.
\]

Conversely, an every-visit MC-control algorithm can use UCB as its
action-selection rule.

Therefore,
\[
\boxed{
\text{UCB/UCT specifies exploration; it does not define the MC update}.
}
\]

Different action-selection rules may have very different exploration
behavior and efficiency, but they operate within the same basic
trajectory-sampling and Monte Carlo updating framework.

\section{The Four MCTS Stages in MC-Control Language}

We can now translate the standard four-stage description directly.

\begin{center}
\begin{tabular}{p{0.28\textwidth}p{0.58\textwidth}}
\toprule
\textbf{MCTS terminology}
&
\textbf{MC-control interpretation}
\\
\midrule

Selection
&
Sample actions using the learned part of the current policy
\\

Expansion
&
First visit and initialization of a previously unrepresented
state/action
\\

Simulation / rollout
&
Continue sampling using the initial part of the current policy
\\

Backup
&
Every-visit Monte Carlo return update
\\

Tree policy
&
Current policy on states with learned value information
\\

Rollout policy
&
Initial/default part of the current policy on states not yet learned
\\

$N(s,a)$
&
Visit count
\\

$Q(s,a)$
&
Empirical Monte Carlo action-value estimate
\\

UCT/UCB
&
One possible exploration/action-selection rule
\\

Search tree
&
Representation of the visited region
\\

\bottomrule
\end{tabular}
\end{center}

The conventional four stages can therefore be reduced to two basic
operations:
\[
\boxed{
\underbrace{
\text{Selection + Expansion + Simulation}
}_{\text{trajectory sampling}}
}
\]
followed by
\[
\boxed{
\underbrace{
\text{Backup}
}_{\text{every-visit MC updating}}.
}
\]

This is the central observation of the note.

\section{Planning Versus Learning}

Another reason MCTS and MC control appear different is that MCTS is
usually called \emph{planning}, whereas MC control is usually called
\emph{learning}.

This distinction is useful at the application level.

MC control is often described as learning values or a policy over a
substantial part of the state space.

MCTS usually starts from a particular current state $s_0$ and
concentrates computation on states relevant to the decision at $s_0$.

Thus, a useful informal distinction is
\[
\text{broad/global Monte Carlo control}
\]
versus
\[
\text{root-localized Monte Carlo control}.
\]

This difference is computationally important. Concentrating simulations
on the states relevant to the current decision is one of the main
reasons MCTS is useful in very large decision spaces.

But localization does not change the underlying Monte Carlo operation:
trajectories are sampled and their returns are used to update the
visited action values.

Planning and learning therefore describe different uses and scopes of
the computation without necessarily implying different underlying
Monte Carlo control mechanisms.

\section{Why Do They Look Like Different Algorithms?}

The distinction is partly historical.

MC control developed within dynamic programming, stochastic control,
and reinforcement learning.

MCTS became prominent within planning, search, and computer games
\cite{browne2012survey}.

The two traditions use different vocabularies.

Where MC control says
\[
\text{sample an action from the current policy},
\]
MCTS distinguishes between
\[
\text{tree policy}
\quad\text{and}\quad
\text{rollout policy}.
\]

Where MC control says
\[
\text{first visit},
\]
MCTS says
\[
\text{expansion}.
\]

Where MC control says
\[
\text{update the action value using the sampled return},
\]
MCTS says
\[
\text{backup}.
\]

These terms are useful. In particular, they make the computational
structure of a search procedure easy to describe and visualize.

But different terminology does not necessarily imply a different
underlying stochastic mechanism.

\section{A Note on the Existing Unified View}

The relationship between MCTS and reinforcement learning has been
studied explicitly before.

Vodopivec, Samothrakis, and \v{S}ter
\cite{vodopivec2017mcts} provide a detailed analysis of this
relationship and argue for greater cross-awareness between the MCTS
and RL communities. Their treatment places MCTS within a broader
family of reinforcement-learning and planning methods and shows how
RL semantics can be used to interpret and extend tree-search
algorithms.

The present note is complementary in purpose. Rather than developing
a general taxonomy of MCTS and RL algorithms, we focus on one
elementary correspondence:

\[
\boxed{
\text{MCTS trajectory generation}
+
\text{MCTS backup}
}
\]
can be read directly as
\[
\boxed{
\text{MC-control trajectory sampling}
+
\text{every-visit MC updating}.
}
\]

This simple viewpoint is useful because it removes much of the apparent
conceptual distance between the two methods without requiring any new
algorithmic machinery.

\section{Implications for Teaching}

The equivalence provides a particularly simple way to introduce MCTS
after Monte Carlo control.

Once every-visit MC control has been taught, begin with
\[
Q(s,a)
\leftarrow
Q(s,a)
+
\frac{1}{N(s,a)}
\bigl(G-Q(s,a)\bigr).
\]

Use the learned $Q$ values and visit information to modify action
selection.

At states that have not yet been learned, retain the initial policy:
\[
\pi_n(\cdot\mid s)=\pi_0(\cdot\mid s).
\]

Create state--action statistics only when they are first needed.

Finally, concentrate the simulations around the current decision
state.

The resulting procedure is what is conventionally described as MCTS.

Under this presentation, the four MCTS stages are not four new
learning ideas. They are a useful operational decomposition of
trajectory sampling and Monte Carlo updating.

The search tree can then be introduced as a natural visualization of
which parts of the policy have already been informed by simulation and
which parts still use their initial values.

\section{Discussion}

Viewing MCTS as every-visit MC control does not diminish the practical
importance of MCTS.

Its success comes from combining a simple Monte Carlo mechanism with
effective computational organization.

Among the practically important ideas are:

\begin{itemize}
    \item concentrating computation around the current decision;
    \item expanding only portions of a very large decision space that
          become relevant;
    \item allocating simulations adaptively among competing actions;
    \item using effective exploration rules such as UCB;
    \item incorporating domain knowledge or learned priors into action
          selection and rollout.
\end{itemize}

Modern systems may further combine search with learned policy and
value functions.

These developments can dramatically change computational efficiency
and decision quality.

The point of the present note is narrower: beneath these choices, the
basic value-learning loop remains
\[
\boxed{
\text{sample}
\rightarrow
\text{return}
\rightarrow
Q\text{-update}
\rightarrow
\text{modified future sampling}.
}
\]

That is precisely the structure of Monte Carlo control.

\section{Conclusion}

This note asked a simple question:

\begin{center}
\emph{Is Monte Carlo Tree Search just every-visit Monte Carlo control?}
\end{center}

At the level of trajectory sampling and Monte Carlo action-value
updating considered here, the answer is yes.

The current policy can be written as
\[
\pi_n(a\mid s)
=
\begin{cases}
\Phi(Q_n(s,\cdot),N_n(s,\cdot)),
&
s\in\mathcal V_n,
\\[2mm]
\pi_0(a\mid s),
&
s\notin\mathcal V_n.
\end{cases}
\]

MCTS calls the first part the tree policy and the second part the
rollout policy.

They can instead be viewed simply as the learned and not-yet-learned
parts of one evolving policy.

Likewise,

\[
\text{Expansion}
=
\text{first visit / initialization},
\]

and

\[
\text{Backup}
=
\text{every-visit Monte Carlo updating}.
\]

UCT is one possible exploration rule, and the search tree is a useful
representation of the visited region.

Consequently,
\[
\boxed{
\text{Selection + Expansion + Simulation}
=
\text{trajectory sampling},
}
\]
and
\[
\boxed{
\text{Backup}
=
\text{every-visit MC updating}.
}
\]

Thus, when stripped of differences in terminology, representation, and
computational emphasis,
\[
\boxed{
\text{Monte Carlo Tree Search}
=
\text{Every-Visit Monte Carlo Control}
}
\]
in the sense described in this note.

The two names remain useful because they emphasize different
computational contexts. But recognizing their common underlying
mechanism provides a simpler way to understand, teach, and relate the
two methods.

\bibliographystyle{plainnat}
\bibliography{references}

@book{sutton2018reinforcement,
  author    = {Richard S. Sutton and Andrew G. Barto},
  title     = {Reinforcement Learning: An Introduction},
  edition   = {2},
  publisher = {MIT Press},
  address   = {Cambridge, MA},
  year      = {2018}
}

@incollection{coulom2006efficient,
  author    = {R{\'e}mi Coulom},
  title     = {Efficient Selectivity and Backup Operators in Monte-Carlo Tree Search},
  booktitle = {Computers and Games},
  pages     = {72--83},
  publisher = {Springer},
  year      = {2006},
  doi       = {10.1007/978-3-540-75538-8_7}
}

@inproceedings{kocsis2006bandit,
  author    = {Levente Kocsis and Csaba Szepesv{\'a}ri},
  title     = {Bandit Based Monte-Carlo Planning},
  booktitle = {Machine Learning: ECML 2006},
  series    = {Lecture Notes in Computer Science},
  volume    = {4212},
  pages     = {282--293},
  publisher = {Springer},
  year      = {2006},
  doi       = {10.1007/11871842_29}
}

@article{browne2012survey,
  author  = {Cameron B. Browne and Edward Powley and Daniel Whitehouse
             and Simon M. Lucas and Peter I. Cowling and
             Philipp Rohlfshagen and Stephen Tavener and Diego Perez
             and Spyridon Samothrakis and Simon Colton},
  title   = {A Survey of Monte Carlo Tree Search Methods},
  journal = {IEEE Transactions on Computational Intelligence and AI in Games},
  volume  = {4},
  number  = {1},
  pages   = {1--43},
  year    = {2012},
  doi     = {10.1109/TCIAIG.2012.2186810}
}

@article{chang2005adaptive,
  author  = {Hyeong Soo Chang and Michael C. Fu and Jiaqiao Hu
             and Steven I. Marcus},
  title   = {An Adaptive Sampling Algorithm for Solving Markov Decision Processes},
  journal = {Operations Research},
  volume  = {53},
  number  = {1},
  pages   = {126--139},
  year    = {2005},
  doi     = {10.1287/opre.1040.0145}
}

@article{vodopivec2017mcts,
  author  = {Tom Vodopivec and Spyridon Samothrakis and Branko {\v S}ter},
  title   = {On Monte Carlo Tree Search and Reinforcement Learning},
  journal = {Journal of Artificial Intelligence Research},
  volume  = {60},
  pages   = {881--936},
  year    = {2017},
  doi     = {10.1613/jair.5507}
}

@inproceedings{fu2018mcts,
  author    = {Michael C. Fu},
  title     = {Monte Carlo Tree Search: A Tutorial},
  booktitle = {Proceedings of the 2018 Winter Simulation Conference},
  pages     = {222--236},
  year      = {2018},
  publisher = {IEEE},
  doi       = {10.1109/WSC.2018.8632344}
}

\end{document}